\documentclass{article} %
\usepackage{iclr2026_conference,times}
\usepackage{graphicx}
\usepackage{listings}
\usepackage{float} %

\usepackage{amsmath,amsfonts,bm}

\def\eqref#1{equation~\ref{#1}}

\def\1{\bm{1}}

\DeclareMathAlphabet{\mathsfit}{\encodingdefault}{\sfdefault}{m}{sl}
\SetMathAlphabet{\mathsfit}{bold}{\encodingdefault}{\sfdefault}{bx}{n}

\usepackage[utf8]{inputenc} %
\usepackage[T1]{fontenc}    %
\usepackage{hyperref}       %
\usepackage{url}            %
\usepackage{booktabs}       %
\usepackage{amsfonts}       %
\usepackage{nicefrac}       %
\usepackage{microtype}      %
\usepackage{xcolor}         %
\usepackage{standalone}
\usepackage[utf8]{inputenc}
\usepackage[T1]{fontenc}

\usepackage{latexsym}
\usepackage{amsmath}
\usepackage{amssymb}
\usepackage{amsthm}
\usepackage{natbib}   
\usepackage{graphicx}
\usepackage{subcaption}
\usepackage{array}
\usepackage{tabu}
\usepackage{makecell}
\usepackage{paralist}
\usepackage{cases}
\usepackage{diagbox}
\usepackage{enumitem}
\usepackage{soul}
\usepackage{multirow}
\usepackage{verbatim}
\usepackage{tabulary}
\usepackage{booktabs}
\usepackage{tabularx}
\usepackage[mathscr]{euscript}
\usepackage{mathtools}
\usepackage{algorithm}
\usepackage{algpseudocode}
\usepackage{stmaryrd}
\usepackage{tikz-dependency}
\usetikzlibrary{automata,decorations.markings,arrows,positioning,matrix,calc,patterns,angles,quotes,calc}
\usepackage{adjustbox}
\usepackage{tabularx}
\usepackage{xspace}
\usepackage{tabulary}
\usepackage{afterpage}
\usepackage{bm}
\usepackage{color}
\usepackage{graphicx}
\usepackage{slashbox}
\usepackage[toc,page]{appendix}
\usepackage{makecell}
\usepackage{boldline}
\usepackage[shortcuts]{extdash}  %

\usepackage{blindtext}
\usepackage{graphicx}
\usepackage{capt-of}
\usepackage{booktabs}
\usepackage{varwidth}
\usepackage{pifont}%
\usepackage{wrapfig}

\usepackage{listings}
\usepackage{pythonhighlight}

\usepackage[normalem]{ulem}
\useunder{\uline}{\ul}{}
\usepackage{ragged2e} %

\usepackage{comment} %

\usepackage{needspace}  %

\definecolor{orange}{rgb}{1,0.5,0}
\definecolor{mdgreen}{rgb}{0.05,0.6,0.05}
\definecolor{mdblue}{rgb}{0,0,0.7}
\definecolor{dkblue}{rgb}{0,0,0.5}
\definecolor{dkgray}{rgb}{0.3,0.3,0.3}
\definecolor{slate}{rgb}{0.25,0.25,0.4}
\definecolor{gray}{rgb}{0.5,0.5,0.5}
\definecolor{ltgray}{rgb}{0.7,0.7,0.7}
\definecolor{purple}{rgb}{0.7,0,1.0}
\definecolor{lavender}{rgb}{0.65,0.55,1.0}

\definecolor{mypurple}{RGB}{111,61,121}
\definecolor{myblue}{RGB}{46,88,180}
\definecolor{myred}{RGB}{181,68,106}
\definecolor{myyellow}{RGB}{204,143,55}

\newcommand{\ensuretext}[1]{#1}
\newcommand{\marker}[2]{\ensuremath{^{\textsc{#1}}_{\textsc{#2}}}}
\newcommand{\draftcomment}[3]{\ensuretext{\textcolor{#3}{[#1 #2]}}}
\renewcommand{\draftcomment}[3]{}  %

\newif\ifqiruncomments
\qiruncommentstrue    %

\newcommand{\term}[1]{\textit{#1}} %

\DeclareSymbolFont{extraup}{U}{zavm}{m}{n}
\DeclareMathSymbol{\vardiamond}{\mathalpha}{extraup}{87}

\newcolumntype{L}[1]{>{\raggedright\let\newline\\\arraybackslash\hspace{0pt}}m{#1}}
\newcolumntype{C}[1]{>{\centering\let\newline\\\arraybackslash\hspace{0pt}}m{#1}}
\newcolumntype{R}[1]{>{\raggedleft\let\newline\\\arraybackslash\hspace{0pt}}m{#1}}

\theoremstyle{definition}

\theoremstyle{remark}

\algrenewcommand{\algorithmiccomment}[1]{\leavevmode$\triangleright$ #1}

\setul{1pt}{.4pt}

\DeclareFixedFont{\ttb}{T1}{txtt}{bx}{n}{12} %
\DeclareFixedFont{\ttm}{T1}{txtt}{m}{n}{12}  %

\newcommand{\cf}{counterfactual\xspace}
\newcommand{\interv}{interventional\xspace}

\setlist[enumerate]{leftmargin=*, itemsep=1mm, parsep=1mm, topsep=1mm, partopsep=1mm}
\setlist[itemize]{leftmargin=*, itemsep=1mm, parsep=1mm, topsep=1mm, partopsep=1mm}

\colorlet{punct}{red!60!black}
\definecolor{background}{HTML}{EEEEEE}
\definecolor{delim}{RGB}{20,105,176}
\colorlet{numb}{magenta!60!black}

\lstdefinelanguage{json}{
    basicstyle=\normalfont\ttfamily,
    numbers=left,
    numberstyle=\scriptsize,
    stepnumber=1,
    numbersep=8pt,
    showstringspaces=false,
    breaklines=true,
    frame=lines,
    backgroundcolor=\color{background},
    literate=
     *{0}{{{\color{numb}0}}}{1}
      {1}{{{\color{numb}1}}}{1}
      {2}{{{\color{numb}2}}}{1}
      {3}{{{\color{numb}3}}}{1}
      {4}{{{\color{numb}4}}}{1}
      {5}{{{\color{numb}5}}}{1}
      {6}{{{\color{numb}6}}}{1}
      {7}{{{\color{numb}7}}}{1}
      {8}{{{\color{numb}8}}}{1}
      {9}{{{\color{numb}9}}}{1}
      {:}{{{\color{punct}{:}}}}{1}
      {,}{{{\color{punct}{,}}}}{1}
      {\{}{{{\color{delim}{\{}}}}{1}
      {\}}{{{\color{delim}{\}}}}}{1}
      {[}{{{\color{delim}{[}}}}{1}
      {]}{{{\color{delim}{]}}}}{1},
}

\newcommand{\para}[1]{\noindent\textbf{#1}}

\title{Executable Counterfactuals: Improving LLMs' causal reasoning through code}

\author{%
  Aniket Vashishtha\,$^1$\thanks{Equal contribution.} 
  \quad
  Qirun Dai\,$^{2*}$ 
  \quad
  \textbf{Hongyuan Mei}\,$^3$ 
  \\
  \textbf{Amit Sharma}\,$^4$\thanks{Equal advising.
  \\ Correspondence to: aniketv2@illinois.edu, qirundai@uchicago.edu 
  } 
  \quad
  \textbf{Chenhao Tan}\,$^{2\dagger}$ 
  \quad
  \textbf{Hao Peng}\,$^{1\dagger}$ 
  \\
  $^1$University of Illinois Urbana-Champaign 
  \quad
  $^2$The University of Chicago\\
  $^3$TTIC 
  \quad
  $^4$Microsoft Research India\\
}

\iclrfinalcopy %
\begin{document}

\maketitle

\begin{abstract}
Counterfactual reasoning, a hallmark of intelligence, consists of three steps: 
inferring latent variables from observations (\term{abduction}), constructing alternative situations (\term{intervention}), and predicting the outcomes of the alternatives (\term{prediction}). 
This skill is essential for advancing LLMs' causal understanding and expanding their applications in high-stakes domains such as scientific research and healthcare. 
However, existing efforts in assessing LLM's counterfactual reasoning capabilities tend to skip the abduction step, effectively reducing to interventional reasoning and leading to over-estimated LLM performance.
To address this, we introduce \textit{executable counterfactuals}, a novel framework that operationalizes causal reasoning through code and math problems. 
Our framework explicitly requires all three steps of counterfactual reasoning and enables scalable synthetic data creation with varying difficulty, creating a new frontier for evaluating and improving LLM's reasoning. 
Our results reveal substantial drop in accuracy (25-40\%) from interventional to counterfactual reasoning for state-of-the-art models such as \textit{o4-mini} and \textit{Claude-4-Sonnet}. %
To address this gap, we construct a training set comprising counterfactual code problems having \textit{if-else condition} and test on out-of-distribution code structures (e.g.,  having \textit{while-loop}); we also test whether a model trained on code can generalize to counterfactual math word problems.
While supervised finetuning (SFT) on stronger models' reasoning traces improves in-distribution performance of Qwen models, it leads to a \textit{decrease }in accuracy on out-of-distribution tasks.
In contrast, reinforcement learning (RL) induces the core cognitive behaviors and generalizes to new distributions, yielding substantial accuracy gains over the base model %
on both code ($\uparrow$\textasciitilde1.5\text{--}2X)  and counterfactual math problems. %
Analysis of the reasoning traces further reinforces these findings and highlights the promise of RL with scalable data generation for improving LLMs' counterfactual reasoning.
Our code and data are available at ~\url{https://github.com/AniketVashishtha/Executable_Counterfactuals}.

\end{abstract}

\section{Introduction}

Counterfactual reasoning is the cognitive process of answering \textit{what-if} questions that underpin critical domains such as scientific discovery \citep{Scholkopf2021}, healthcare \citep{Richens2020CausalDx}, economics \citep{AtheyImbens2017JEP}, and public policy \citep{PoulosZeng2021RNNcf}. Given an action and an observed outcome, it involves  
inferring the latent state of a system when the action was performed (\term{abduction}), constructing alternative scenarios through \term{interventions}, and \term{predicting} the outcomes under those counterfactual scenarios~\citep{Pearl_2002, EpstudeRoese2008FunctionalCFT}. %
Despite the importance of counterfactual reasoning, it remains a widely documented weakness of current large language models (LLMs;~\citealp{jin2024cladderassessingcausalreasoning, yamin2025llmsstruggleperformcounterfactual, yu2023ifqadatasetopendomainquestion}). 

\begin{figure}[t]
  \centering
  \includegraphics[width=\textwidth,trim={206pt 115pt 206pt 115pt},clip]{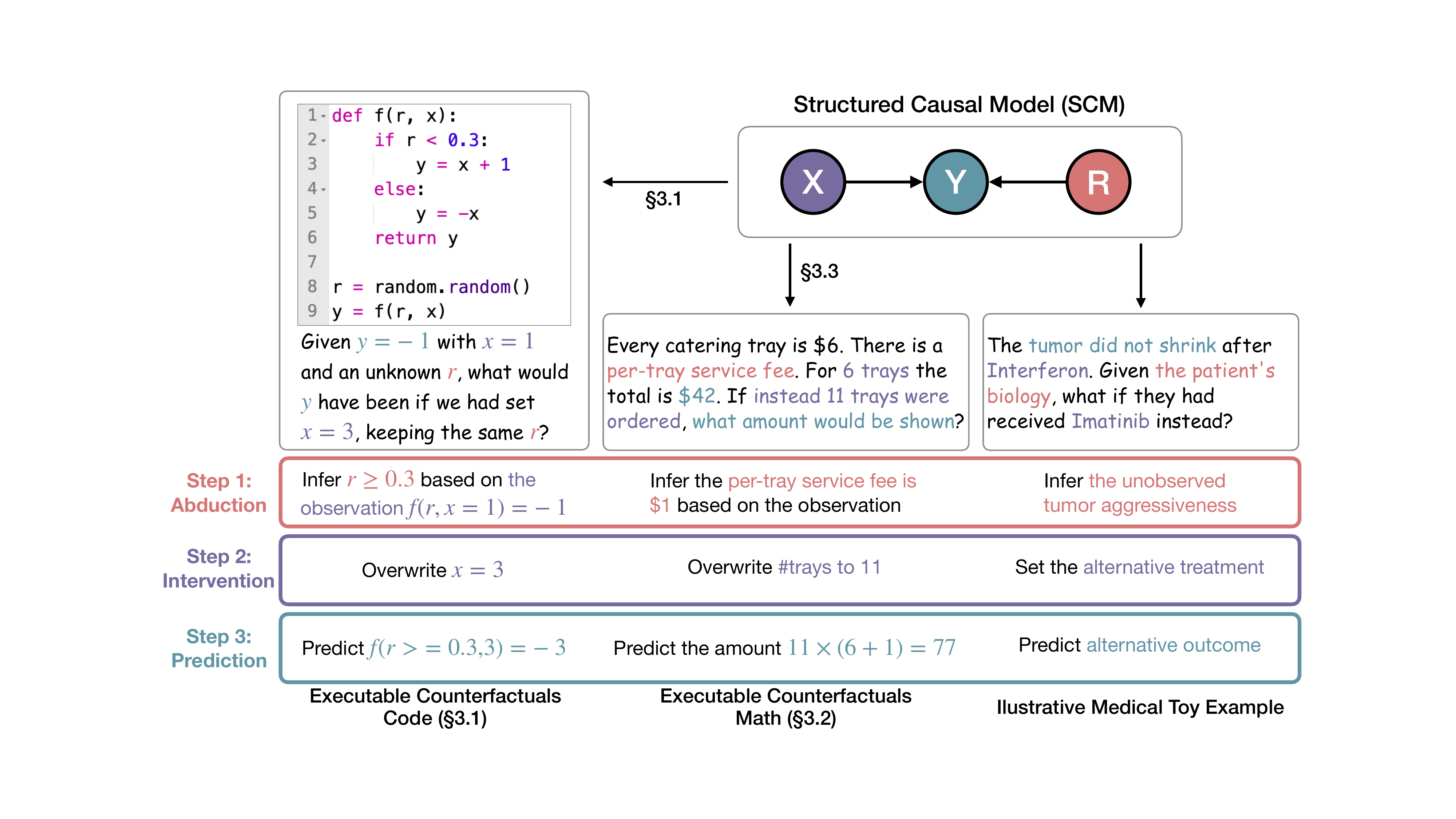}
  \caption{Executable counterfactuals across code, math, and a medical toy example to illustrate abduction--intervention--prediction. Code offers a controlled, executable setting that maps naturally to causal/computational graphs and transfers to natural-language tasks.}
  \label{fig:cf_vs_interv_pipeline}
\end{figure}

Evaluating and improving counterfactual reasoning is challenging because counterfactuals are inherently unobservable and rely on hypothetical alternatives to reality. As a result, prior work  considers either synthetic graph-based settings that are hard to map to real-world problem solving~\citep{jin2024cladderassessingcausalreasoning} or  simplistic tasks that are \textit{expressed} in counterfactual language but can be solved without invoking all aspects of counterfactual reasoning. Examples include binary classification tasks given full information about the causal graph (\textit{Would $Y$ still occur if $X$ didn't happen?}; \citealp{chen2025counterbenchbenchmarkcounterfactualsreasoning}) or benchmarks based on perturbations of existing reasoning problems (thus creating new ``counterfactual'' problems;~\citealp{wu-etal-2024-reasoning}). 
With full information (i.e., there are no latent confounders or noise), these problems can be solved by simple forward reasoning~\citep{gerstenberg2022would}: change the input variables' values as instructed and solve it as a new problem, \emph{without} any counterfactual reasoning.

These simplified interpretations of counterfactuals risk conflating them with simpler forms of causal reasoning (more on this in \S\ref{sec:background})
and thus misrepresent LLMs' counterfactual abilities.
To address these limitations, we identify the three core cognitive skills  from Pearl's definition of counterfactual reasoning~\citep{Pearl_2002}---\term{Abduction} , \term{Intervention}  and \term{Prediction}---and construct tasks that requires all three skills to obtain a correct solution. In line with prior work \citep{gandhi2025cognitivebehaviorsenableselfimproving} that evaluates cognitive behaviors in LLMs for self-improvement, we assess the behaviors required for counterfactual reasoning.
Key benefits of this perspective include explicit separation of counterfactual reasoning from simpler forms of causal reasoning, fine-grained attribution of models' strengths and weaknesses, and an actionable framework for improvement (\S\ref{sec:framework_design}).
Moreover, beyond counterfactuals, improvements to these cognitive skills can independently serve as building blocks for stronger LLM reasoning in general.

Our key idea is to use code understanding as a problem setting for studying counterfactual reasoning (\textit{executable counterfactuals}). We show how  real-world partial information settings can be abstracted in code through latent variables while still allowing for objective evaluation. Specifically, we introduce random variables in the code understanding task such that their values are not revealed to the language model. In the formal structural causal model framework, these random variables can be considered as noise variables that need to be inferred before making any counterfactual prediction. 
As shown in the illustrative example in Figure~\ref{fig:cf_vs_interv_pipeline},
the  causal structure $X\rightarrow Y\leftarrow R$ where $X$ and $R$  independently cause  $Y$,
converts to a program where
$X$ computes $Y$ while $R$ determines conditional branching.
A counterfactual question is constructed as:
\textit{Given observation $y = f(r, x=1) = -1$ with unknown $r$, what would $y$ have been if we had set $x = 3$, keeping the same $r$?}
Solving this problem requries the agent to invoke all three cognitive skills, 
(1) infer $r$ based on the observation $y=-1$ (abduction),
(2) mentally set $x=3$ (intervention),
and (3) compute the resulting $y$ (prediction).

Beyond the aforementioned benefits,
our code-based framework
avoids the potential ambiguity of natural language,
and allows rich and controllable complexity for constructing evaluation problems and generating synthetic training data (\S\ref{sec:framework_design}).
It evaluates models' ability to use counterfactual reasoning for problem solving rather than reducing the task to answer binary classification questions~\citep{jin2024cladderassessingcausalreasoning,chen2025counterbenchbenchmarkcounterfactualsreasoning}.
In addition,  it facilitates evaluating and improving out-of-distribution generalization by varying the program structures and translating coding tasks into counterfactual math problems \S\ref{sec:framework_design}. We address the following important research questions\label{sec:rqs} with executable counterfactuals:
\begin{enumerate}[noitemsep,topsep=0pt,parsep=0pt,partopsep=0pt]
    \item \textbf{\textit{How do current LLMs perform on counterfactual reasoning?}} 
    Our experiments with open models of sizes ranging from 1.5B to 72B parameter and commercial reasoning models show strong performance on straightforward code-execution tasks, but poor performance on counterfactual reasoning over the same code.
    Qualitative analysis indicates consistent failure at the abduction step, leading to incorrect conditioning on the original observation to infer latent features.
    
    \item \textbf{\textit{Can SFT distillation from stronger models instill these skills and do they generalize?}} 
    We finetune Qwen 1.5B/3B/7B-Instruct on reasoning trajectories from DeepSeek-Distilled-Qwen-32B,
    and observe \textasciitilde$40\%$ 
    performance improvements on in-domain evaluation.
    However, these improvements do not generalize to unseen code structures or counterfactual math problems, highlighting the limited generalization of SFT.
    
    \item \textbf{\textit{How does RL fare?}} 
    Training the same models with RL from verifiable rewards (RLVR) 
    using GRPO~\citep{shao2024deepseekmath} 
    leads the models to acquire the necessary cognitive skills, showing strong transfer across diverse code structures and counterfactual math problems in natural language, with concrete evidence of improved generalization.
\end{enumerate}

Our findings have two key implications. 
First, they reinforce recent evidence that current LLMs remain weak at counterfactual and causal reasoning \citep{jin2024cladderassessingcausalreasoning, jin2023can, willig2023probing}.
Second, our experiments call into question the effectiveness of SFT, 
a widely adopted approach by recent works to improve counterfactual reasoning \citep{guo2025deepseek, li2025naturalthoughts},
especially regarding its ability to generalize to complex and high-impact real-world domains.
In contrast, our results show that RL elicits stronger generalization for counterfactual reasoning; despite training only on code, the model internalizes the core skills and applies them directly to counterfactual math problems, providing early evidence that RL is a promising pathway for eliciting such reasoning in LLMs.
Crucially, as  shown in the experiments, our code-based framework has the potential to offer a scalable way for 
learning counterfactual reasoning that transfers to new domains where training data can be scarce.
All code and data will be publicly released upon publication.%

\section{Background and Related Work}
\label{sec:background}

In this section, we will first outline the cognitive skills required for counterfactual reasoning 
and then show how it is often conflated with interventional reasoning in prior work.

\para{From abduction to prediction.}
We use Figure~\ref{fig:cf_vs_interv_pipeline} as a running example to expand the cognitive skills required for counterfactual reasoning.
Three steps are needed to answer the counterfactual question
\textit{Given observation $y=f(r, x = 1)$, what would  \(y\) have been had \(x=3\) instead in the original run?}

\begin{itemize}[noitemsep,topsep=0pt,parsep=0pt,partopsep=0pt]
    \item[] \textbf{Step 1: Hindsight reasoning for abduction}: 
    Rewind back to the point where the original action was taken, to infer latent features and noise present in the system at that time. 
    The above counterfactual question,
    cannot be answered by simply re-running the program with \(x=3\). One must first \emph{abduce} the hidden latent variable \(\hat r\geq 0.3\) from the observed run \(f(\hat r, x=1)=1\).

    \item[] \textbf{Step 2: Taking a different action (intervention)}: Conditioned on the inferred latent features from abduction stage, perform the counterfactual change by intervening the input to its counterfactual value while keeping everything else the same as in the earlier observation.
    For the code example, this means holding \(\hat r\) fixed while intervening by overwriting \(x=3\).
    
    \item[] \textbf{Step 3: Prediction}: Based on the new action taken, compute its consequences in the counterfactual scenario. In the example, computing \(y_{\mathrm{cf}} = f(\hat r\geq0.3, x=3)\) is final prediction step.
\end{itemize}

Without latent states and the abduction step, counterfactual reasoning reduces to interventional reasoning, corresponding to Level 2 in Pearl's causal ladder~\citep{pearl2009causality}, which breaks down causal reasoning to
three progressively more advanced levels: Associational (Level 1), Interventional (Level 2), and Counterfactual (Level 3); see Appendix~\ref{sec:ladder} for a detailed overview.

\para{Past studies often overlook abduction.}
Prior evaluations of LLM counterfactual reasoning often use fully observed settings with no latent noise.
This effectively makes the abduction step unnecessary since there is \emph{no} unobserved variable or noise to abduce. 
In such regimes, a \textit{counterfactual} query collapses to an \textit{interventional} one: the answer follows directly from taking a different action not requiring the step of inferring hidden state. 
Take Figure~\ref{fig:cf_vs_interv_pipeline} (left) as an example and consider the following question $q$:
\textit{What would $y$ have been be if we had set $r=0.4, x=3$?}
Although $q$ may appear similar to the counterfactual question in Figure~\ref{fig:cf_vs_interv_pipeline}, it is fundamentally different.
Crucially, answering $q$ does not require abducting the values of $r$, since it is explicitly specified.
Therefore, solving $q$ relies solely on interventional reasoning (Level 2) rather than counterfactual reasoning (Level 3); in this sense, $q$ effectively collapses to an interventional question despite its seemingly ``counterfactual'' framing.

The above example question $q$, though synthetic, conceptually illustrates the key reason for the mischaracterization of counterfactuals in many recent works \citep{wu-etal-2024-reasoning,li2024promptinglargelanguagemodels, chen2025counterbenchbenchmarkcounterfactualsreasoning, nguyen2024llmsgeneratingevaluatingcounterfactuals, paranjape-etal-2022-retrieval, wu-etal-2021-polyjuice, madaan2021generatecounterfactualscontrolledcounterfactual, ye-etal-2021-connecting, joshi2022investigationineffectivenesscounterfactuallyaugmented, vashishtha2023evaluatingmitigatinggenderbiases}.\footnote{
It should be acknowledged that many of these works focus on
robustness, generalization, and debiasing, and never intend to study counterfactuals as in the causal sense.
Nonetheless, the loose use of the counterfactual framing can lead to misinterpretations by the readers \cite{zhao-etal-2018-gender, kaushik2020learningdifferencemakesdifference, vashishtha2023evaluatingmitigatinggenderbiases}, which highlights the importance of a precise characterization of counterfactuals.
}
See Appendix~\ref{sec:interv_same_cf}  for detailed discussion.%

Clearly distinguishing counterfactual from interventional reasoning is important for accurately understanding the capabilities and limitations of current LLM paradigms, and for designing algorithms that advance their causal reasoning.
It requires an explicit characterization of the three-step process of abduction, intervention, and prediction,
which motivates our executable counterfactual framework.

\para{Other related work.}
\citet{jin2024cladderassessingcausalreasoning} provides formal benchmarks across the causal ladder, including counterfactuals. While well grounded in causal theory, some tasks are less aligned with realistic applications and often presuppose familiarity with advanced tools (do-calculus, d-separation, mediation/IV) which can make it harder to pinpoint whether errors stem from graph inference, identifiability, effect decomposition, or numerical estimation. Similar trade-offs appear in recent causal benchmarks \citep{yang2025eligibilityllmscounterfactualreasoning, zhou2024causalbenchcomprehensivebenchmarkcausal, tandon2019wiqadatasetwhatif}.
Operating on code and math, two domains where recent LLMs have made rapid progress, our framework provides concrete mechanisms to isolate their causal capabilities and to apply established methods such as SFT and RLVR to enhance counterfactual reasoning, as we will do in the experiments.

\section{Operationalizing Counterfactual Reasoning via Code \& Math} %
\label{sec:framework_design}

 We move beyond graphical approaches \citep{yang2025eligibilityllmscounterfactualreasoning} and purely formal tests \citep{jin2023can, jin2024cladderassessingcausalreasoning} by using executable code as an actionable environment for counterfactual reasoning. 
Because programs are computational graphs, they map naturally onto mathematical and graph formalisms
and enable fine-grained control of task difficulty and latent-variable structures.
This allows for designing out-of-distribution (OOD) 
evaluation by encoding causal graphs with novel features and logic unseen during training.
Our framework produces executable counterfactuals with verifiable ground-truth outcomes for both evaluation and training.

\subsection{Executable Counterfactuals: Code} \label{sec:code}

\begin{wrapfigure}{r!}{.43\textwidth}   %
\begin{minipage}{\linewidth}
    \vspace{-.5cm}
    \centering\captionsetup[subfigure]{justification=centering}
    \includegraphics[width=\linewidth,trim={675pt 330pt 675pt 330pt},clip]{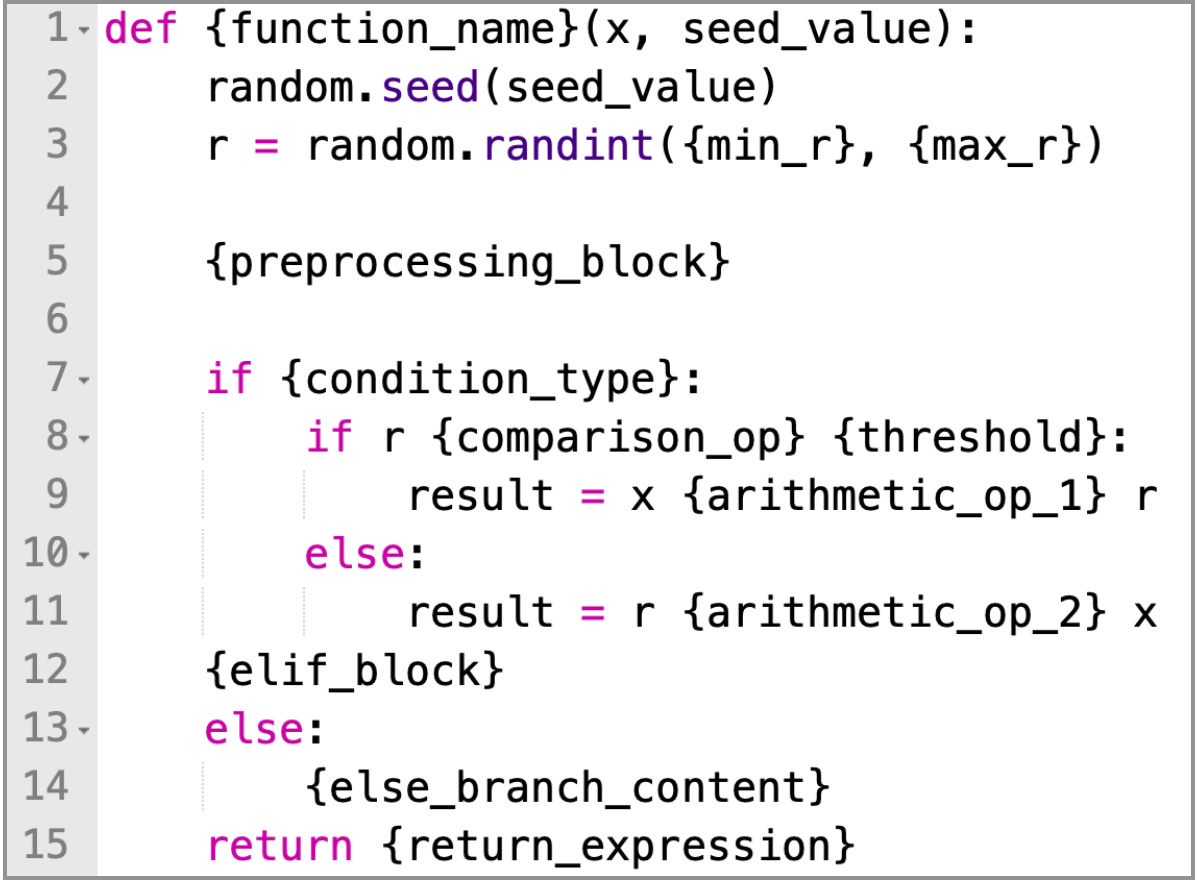}
    \subcaption{Template instance for generating if-else functions in the training set}
    \label{fig:meta_template}\par\vfill%
    \includegraphics[width=\linewidth,trim={675pt 373pt 675pt 373pt},clip]{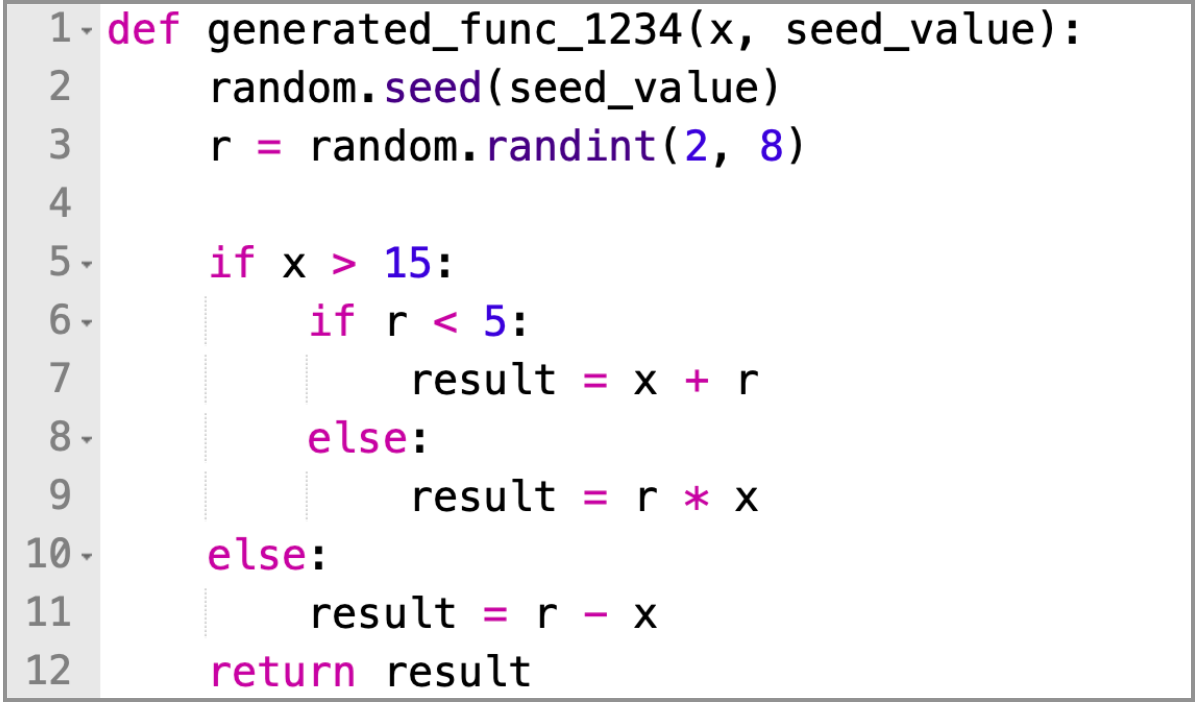}
    \subcaption{Code function generated from template in \ref{fig:meta_template}}
    \label{fig:code1}\par\vfill%
    \includegraphics[width=\linewidth,trim={675pt 359pt 675pt 359pt},clip]{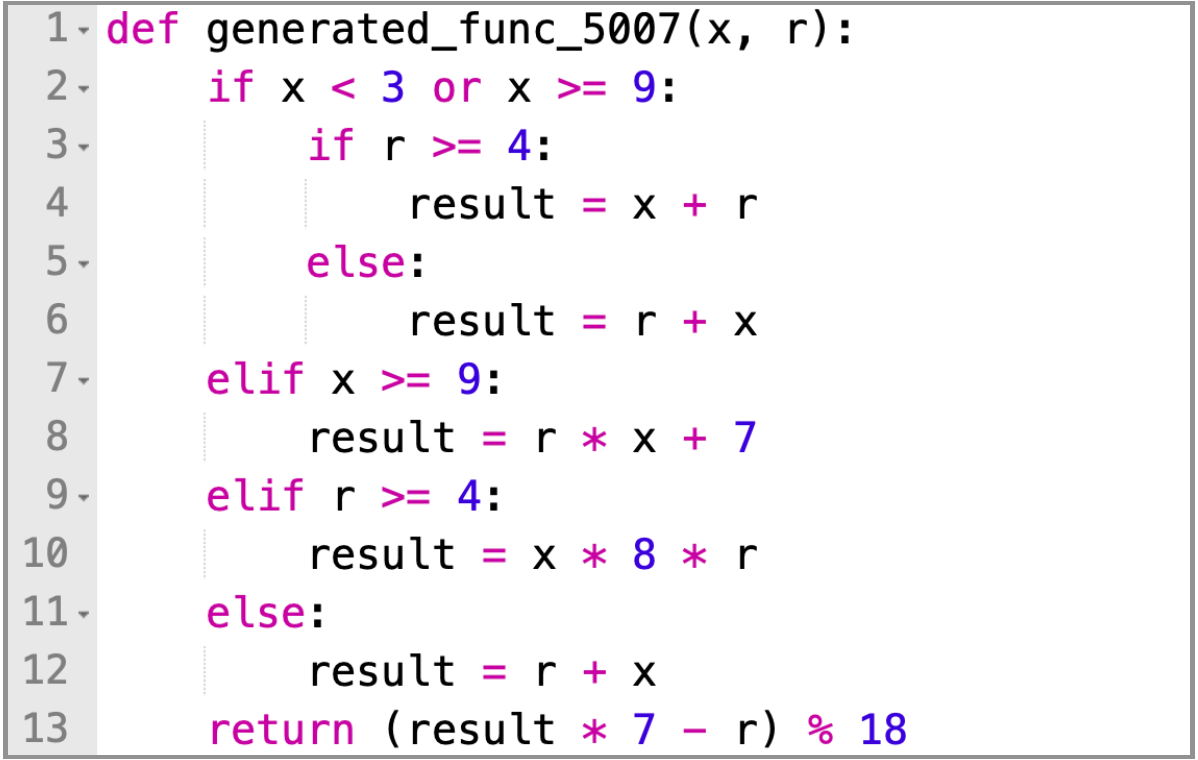}
    \subcaption{Another structurally different code function generated from the same template in \ref{fig:meta_template}}\label{fig:code2}
    \vspace{-.2cm}
\end{minipage}
\caption{Structural diversity emerges from our nested template based approach where a single template can generate structurally and semantically different functions as shown in \ref{fig:code1} and \ref{fig:code2}}\label{fig:template_generation}
\end{wrapfigure}

\paragraph{Overview.}
We generate distinct and executable Python functions from a small set of templates (8 for training, and 3-4 for each evaluation setup) 
by abstracting out the overall program structure and isolating it from specific variables and operators.
Unlike prior work that typically uses a checklist approach which merely swaps numbers or operators while keeping the same control flow~\citep{ribeiro2020accuracybehavioraltestingnlp},
we use function templates where complete code blocks with different functional purposes are replaced by empty placeholders (Figure~\ref{fig:meta_template}).
Specifically, we apply \textit{Claude-4-Sonnet} to draft these templates and 
potential code block candidates for each placeholder,
and perform manual verification to ensure quality and diversity. 
For each type of dataset split (training or evaluation) and control logic (if-else, while loop, etc.), 
we fix a small set of templates along with a list of code block candidates.
For training datasets, we supply 15 combinations of function templates and code block candidates.
Moreover, to promote finer-grained variations in intermediate computations, we also make operators and variables in the functions changeable.
Finally, we deduplicate the generated functions using techniques in Appendix~\ref{sec:deduplication}, which eventually results in a large and diverse set of executable functions using an efficient and controllable recipe.

\paragraph{Template-based generation.}
We consider the following four function logic:
\begin{enumerate}[noitemsep,topsep=0pt,parsep=0pt,partopsep=0pt]
\item \textbf{If\_else:} 
These simple functions have at most one level of nesting structure, thus keeping the intermediate computational steps at a low level (Figure \ref{fig:meta_template}).

\item \textbf{If\_else-long}: 
To test if the models can generalize to longer code structures with more statements, we construct this evaluation dataset with higher levels of nested if-else structures (Table \ref{tab:counterfactual_prompt-diverse_ifelselong}). 

\item \textbf{While}: 
To test how models generalize \cf reasoning to control logic that it has never seen during training, we construct this dataset with \textit{while loops} (Table \ref{tab:counterfactual_prompt-diverse_while}).

\item \textbf{Multi\_r}: 
To test how models generalize to a different causal structure where multiple hidden variables are present, we construct this dataset where each function has three unknown input arguments. Moreover, we level up the complexity by introducing simple \textit{for loops} (Table \ref{tab:counterfactual_prompt-multi_r}) apart from \textit{if-else} statements.

\end{enumerate}

\textbf{If\_else} is used for both training and in-distribution (ID) evaluation, while
\textbf{If\_else-long}, \textbf{While}, and \textbf{Multi\_r} are used for out-of-distribution (OOD) evaluation and never used in the training data.

One important feature of our template approach is that there are three different levels of placeholders whose combinations can greatly advance the diversity of our final datasets.
\begin{itemize}[noitemsep,topsep=0pt,parsep=0pt,partopsep=0pt]
\item \textbf{Fixed placeholders:} boilerplate such as the function name, a reproducible draw of a latent variable \(r\), by setting the random seed, and the final return statement. To design functions with more than one latent variables, we explicitly define placeholders for each extra latent variable.
\item \textbf{Structural placeholders:} Slots for complete code blocks that define the program's logic, including the optional pre-processing steps, the main \texttt{if}-condition (simple or compound), possible \texttt{elif} clauses, code pieces inside each branch, and the form of the return statement.
\item \textbf{Value placeholders:} Specific operators and numbers (e.g., \texttt{+}, \texttt{*}, thresholds) that determine the function's detailed behavior once the structure is chosen.
\end{itemize}

To better mirror real-world ambiguity, where multiple latent configurations can explain the same observation, we insert a modulo at the return statement in training functions (i.e., \(\mathrm{return}\; g(\cdot) \bmod m\)). The modulo's periodicity induces a many-to-one mapping from latent \(r\) to the observed output, so several \(r\) values are consistent with the factual run, yielding multiple valid counterfactual outcomes. At evaluation, we score the model against the full set of valid answers: we report \textit{exact match} (set equality) and an aggregated \textit{F1} that rewards partial coverage of the ground-truth set.

To create the interventional version of the same programming problem, we keep the code unchanged and disclose the realized value(s) of \(r\). Revealing \(r\) removes the abduction step, so the task reduces to re-evaluating the program under a new input \(x\). 
Please refer to Table~\ref{tab:interventional_prompt_multir} for interventional prompt examples.

\clearpage

\subsection{GSM Math Problem Construction for Counterfactual Reasoning} \label{sec:math}

To test whether models can generalize beyond code, we construct a new dataset of counterfactual variants of GSM-8K-style problems. See Figure~\ref{fig:cf_vs_interv_pipeline} (middle) for an illustrative example.
The key idea is to introduce a hidden factor in each problem.
Taking inspiration from \citet{ye2024physicslanguagemodels21}, each problem starts in an everyday setting (office party, school fundraiser, etc.) and is specified by a computational graph that tracks the key quantities (such as counts, unit prices, or fees) and how they combine (sums, percentages, etc.).
Inside
this graph, we introduce one hidden factor that also contributes to the total;
their value is known in the computational graph but not revealed in the narrative.
The hidden factors are simple but varied. Examples include: flat add-on (e.g., an unseen service fee), 
per-item add-on (e.g., an extra fee per tray), and an unknown amount of additional items at a known unit price (e.g., some dessert boxes per tray at \$3 each).
We verbalize the graph into GSM-style word problems.
To increase variety, we use a small set of phrasing templates for different settings (such as office party, fundraiser, etc) and vary both the scenarios and the point where the hidden factor is introduced into the graph.
Ground truth answers are produced by executing the computational graphs, therefore resulting in verifiable answers. For creating an interventional version of the problem, we keep everything exactly the same and reveal the value of the latent variable in the problem statement (Table~\ref{tab:gsm_cf_vs_interv}). To ensure that the latent variable is used in final answer computation, for each problem constructed, we vary the value of the latent variable and see if it leads to change in final answer. If there is no change we regenerate the problem. 

\section{Experiments}\label{sec:experiments}

\begin{figure}[t] %
  \centering
  \includegraphics[width=\textwidth]{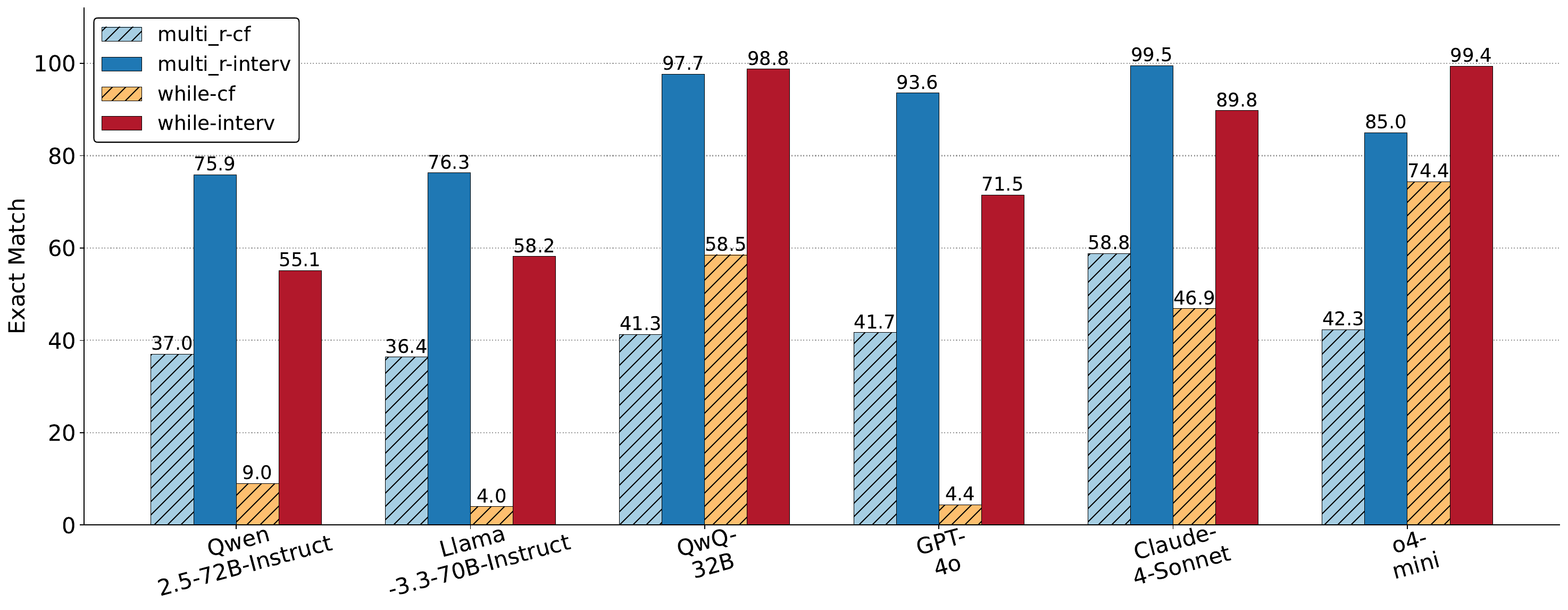}
  \caption{
  Even for LLMs with strong general capabilities or thinking features, the performance gap between \cf and \interv questions originated from the same code function can still be huge, showing the importance of targeted improvements in \cf reasoning.
  }
  \label{fig:code_cf_vs_interv}
\end{figure}

With our executable counterfactuals framework we answer the three research questions  in \S\ref{sec:rqs}.

\subsection{LLMs  Show Weaknesses in Counterfactual Reasoning}

\paragraph{Motivation and Setting.}
As discussed in \S\ref{sec:background}, the lack of abduction in prior works reduces \cf reasoning to 
\interv reasoning, thus failing to 
distinguish the true \cf from \interv capabilities.
In light of this, we pair each of the \cf evaluation dataset of our framework with an \interv counterpart, which is built upon the same code function or mathematical conditions except that the originally hidden variable is now revealed and fixed (Table~\ref{tab:cf-example} and~\ref{tab:interventional_prompt_multir}).%
We evaluate a wide range of models with strong reasoning capabilities and present the comparison results in Figure~\ref{fig:code_cf_vs_interv} and Table~\ref{tab:gsm_cf_vs_interv}. 
Please refer to Appendix~\ref{sec:eval_config} for the 
evaluation hyperparameters adopted throughout this work.

\paragraph{Findings.}
For six strong LLMs spanning four model families in both coding and math domains, there consistently exists a significant performance gap between the \cf datasets of our framework and their \interv counterparts, regardless of model providers, sizes, and test-time scaling features. 
Notably, reasoning models (e.g., QwQ-32B \citep{qwq32b} and o4-mini) show nearly perfect \interv reasoning performance in coding, yet achieve less than half on \cf reasoning. Non-reasoning models mostly score below 10\% in \cf datasets with while loops, but can achieve over 70\% in their \interv counterparts. 
Therefore, our framework reveals the  weakness of current strong LLMs in true \cf reasoning, suggesting the necessity of targeted post-training improvements apart from traditional focus on general capabilities only.

\subsection{Distillation-based SFT Generalizes Poorly}

\begin{table*}[t] %
  \renewcommand{\arraystretch}{1.2} %
  \centering
  \resizebox{\linewidth}{!}{
  \begin{tabular}{@{}>{\centering\arraybackslash}c >{\centering\arraybackslash}c cc m{0.01em} cc m{0.01em} cc m{0.01em} cc@{}}
    \toprule
   \multirow{3}{*}{\textbf{Model Class}} &
   \multirow{3}{*}{\textbf{Model}} &
   \multicolumn{2}{c}{\textbf{ID}} && \multicolumn{8}{c}{\textbf{OOD}}  \\
   \cmidrule{3-4}\cmidrule{6-13}
   & & \multicolumn{2}{c}{\textbf{if\_else}} && %
       \multicolumn{2}{c}{\textbf{if\_else-long}} && %
       \multicolumn{2}{c}{\textbf{multi\_r}} && %
       \multicolumn{2}{c}{\textbf{while}} \\ %
     \cmidrule{3-4}
      \cmidrule{6-7}
       \cmidrule{9-10}
        \cmidrule{12-13}
   & & \bf F1 & \bf EM && \bf F1 & \bf EM && \bf F1 & \bf EM && \bf F1 & \bf EM\\ \midrule

    \multirow{12}{*}{\textbf{\makecell[c]{Controllably\\Trained Models}}} &
Qwen2.5-1.5B-Instruct &
      19.3 & 5.3 &&
      26.5 & 12.8 &&
      9.5 & 7.4 &&
      1.9 & 0.8 \\

& Qwen2.5-1.5B-Instruct-SFT 
    & \textbf{62.7} & \textbf{44.4} &&
      \textbf{51.3} & 32.0 &&
      21.4 & 20.7 &&
      2.8 &  2.4 \\

& Qwen2.5-1.5B-Instruct-RL   
    & 34.7 & 20.2 &&
      50.3 & \textbf{39.6} &&
      \textbf{25.5} & \textbf{25.2} &&
      \textbf{5.0}  & \textbf{4.2} \\
      
    \cmidrule{2-13}

& Qwen2.5-3B-Instruct        
    & 32.1 & 11.8 &&
      38.7 & 16.7 &&
      14.0 & 11.1 &&
      5.4 & 2.7 \\

& Qwen2.5-3B-Instruct-SFT    
    & 70.8 & 53.2 &&
      55.4 & 34.7 &&
      22.8 & 21.6 &&
      2.6 & 2.2 \\

& Qwen2.5-3B-Instruct-RL 
    & \textbf{74.8} & \textbf{55.2} &&
      \textbf{55.9} & \textbf{39.3} &&
      \textbf{36.9} & \textbf{35.9} &&
      \textbf{12.9} & \textbf{10.5} \\
    
    \cmidrule{2-13}

& Qwen2.5-7B-Instruct        
    & 38.8 & 13.9 &&
      54.9 & 28.2 &&
      21.6 & 17.9 &&
      7.3 & 3.3 \\

& Qwen2.5-7B-Instruct-SFT    
    & 75.8 & 59.0 &&
      61.4 & 41.7 &&
      24.9 & 23.3 &&
      2.5 & 2.1 \\

& Qwen2.5-7B-Instruct-RL     
    & \textbf{81.7} & \textbf{67.8} &&
      \textbf{75.0} & \textbf{58.3} &&
      \textbf{40.3} & \textbf{36.3} &&
      \textbf{11.2} & \textbf{8.1} \\
      \midrule
\multirow{5}{*}{\textbf{\makecell[c]{General LLMs}}} 

& Qwen2.5-32B-Instruct 
    & 42.9 & 17.2 &&
      63.3 & 29.9 &&
      40.1 & 34.8 &&
      11.2 & 6.2 \\

& Qwen2.5-72B-Instruct       
    & 47.0 & 20.3 &&
      65.0 & 32.8 &&
      42.3 & 37.0 &&
      13.6 & 9.0 \\

& Llama-3.3-70B-Instruct     
    & 50.0 & 22.0 &&
      62.8 & 28.7 &&
      41.8 & 36.4 &&
      12.0 & 4.0 \\

& GPT-4o                     
    & 50.6 & 25.6 &&
      62.6 & 32.9 &&
      44.8 & 41.7 &&
      10.5 & 4.4 \\

& Claude-4-Sonnet            
    & \textbf{79.1} & \textbf{60.6} &&
      \textbf{81.3} & \textbf{59.0} &&
      \textbf{63.5} & \textbf{58.8} &&
      \textbf{53.0} & \textbf{46.9} \\

     \midrule
    \multirow{3}{*}{\textbf{\makecell[c]{Reasoning LLMs}}} 

& R1-Distill-Qwen-32B 
    & 86.0 & 69.1 &&
      89.7 & 77.9 &&
      \textbf{57.1} & \textbf{47.9} &&
      69.7 & 63.1 \\

& QwQ-32B                    
    & 73.5 & 54.9 &&
      85.1 & 73.0 &&
      44.7 & 41.3 &&
      63.2 & 58.5 \\

& o4-mini                    
    & \textbf{91.1} & \textbf{76.2} &&
      \textbf{95.9} & \textbf{90.2} &&
      51.9 & 42.3 &&
      \textbf{84.6} & \textbf{74.4} \\

      \bottomrule
    \end{tabular}
  }
  \caption{
  Evaluation results on in-distribution (ID) and out-of-distribution (OOD) \cf coding tasks using our executable counterfactuals framework. 
  Since each question may contain multiple answers, we report both F1 and exact match scores in percentage units.
  }
  \label{tab:code_cf-all_results}
\end{table*}

\paragraph{Motivation and Setting.}
We then explore SFT, a widely adopted approach that has been traditionally shown effective for targeted improvements in \cf reasoning~\citep{hüyük2025reasoningelicitationlanguagemodels, huang2024clomocounterfactuallogicalmodification} %
. 
Specifically, we opt for the popular long-Chain-of-Thought (long-CoT)  SFT paradigm, where the CoT annotations are distilled by a reasoning model with thinking features, due to its proved benefits of better transfer in reasoning tasks~\citep{guo2025deepseek, li2025naturalthoughts}. 
We choose \textit{DeepSeek-R1-Distill-Qwen-32B}~\citep{guo2025deepseek} as the teacher model, and \textit{Qwen2.5-1.5B/3B/7B-Instruct} series as the base models for all post-training attempts throughout this work.
Please refer to Appendix~\ref{sec:sft_config} for more data annotation and training details .

\begin{wrapfigure}{r}{0.4\textwidth}
    \centering
    \includegraphics[width=0.4\textwidth]{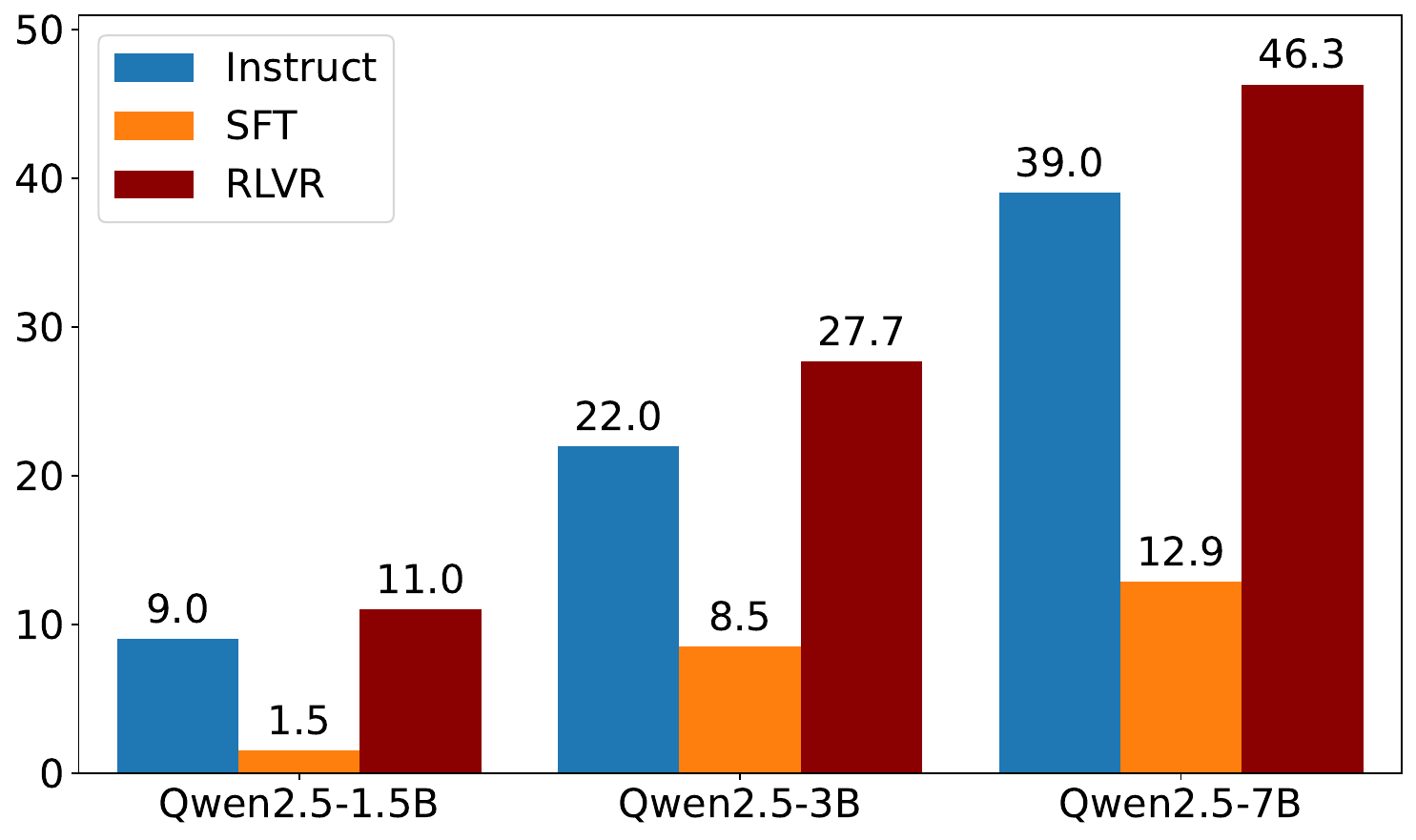}
    \caption{
    Accuracies on the GSM-\cf dataset under domain-transfer. RLVR consistently shows effective generalization from code-based to natural language-based \cf reasoning, while SFT consistently fails. Moreover, the improvements of RLVR also robustly scale with the model size. 
    }
    \label{fig:gsm_cf-inst_sft_rl}
    \vspace{-6mm}   %
\end{wrapfigure}

\paragraph{Findings.}
As shown in Table~\ref{tab:code_cf-all_results} and Figure~\ref{fig:gsm_cf-inst_sft_rl}, compared with their base, SFT models achieve strong in-distribution (ID) \cf reasoning performance, as well as decent performance when certain surface task features (e.g., length of code functions in \textit{if\_else-long}) are out-of-distribution (OOD).
However, when the fundamental reasoning structures of these tasks become OOD, including the \textbf{causal structure} (e.g., more hidden variables in \textit{multi\_r}), \textbf{control logic} (e.g., while loops as the control structure in \textit{while}), and \textbf{question domain} (e.g., from code-based to natural language-based math reasoning in \textit{gsm}), the gains of SFT diminishes and it even hurts the performance in most cases.
Thus, our framework demonstrates that long-CoT SFT paradigm has only limited generalization of \cf reasoning, despite the powerful external supervision signals. These findings call for investigations into other post-training approaches that are not only more supervision-efficient, but can also generalize to complex and previously unseen task structures.

\subsection{RLVR elicits generalizable \cf reasoning skills across causal structures and question domains}

\paragraph{Motivation and Setting.}

In search of a supervision-efficient approach to generalize \cf reasoning capabilities, we eventually resort to reinforcement learning.
We use reinforcement learning from verifiable reward (RLVR) with GRPO~\citep{shao2024deepseekmath},
a popular combination that requires only outcome-based supervision. 
Following prior work~\citep{sun2025omega}, we use exact match scores as the outcome-based reward, and set the prompt batch size and rollout size as 16 and 24   respectively. Please refer to Appendix~\ref{sec:rl_config} for more details about RLVR training.

\paragraph{Findings.}

As shown in Table~\ref{tab:code_cf-all_results} and Figure~\ref{fig:gsm_cf-inst_sft_rl}, RLVR achieves consistent and significant gains for all scales of models, and on all ID and OOD evaluation datasets. The improvements are especially strong on \textit{multi\_r}, \textit{while}, and \textit{gsm}, where involve fundamentally OOD causal structures and reasoning contexts, and make our previous SFT attempt uniformly fail.
Notably, a \textit{Qwen2.5-7B-Instruct} model trained with RLVR achieves comparable performance with \textit{Qwen2.5-72B-Instruct}, and consistently better performance than its 32B variant across the whole coding domain.
Therefore, RLVR successfully achieves our goal of generalizing fundamental \cf reasoning skills to complex structures and previously unseen domains with minimal supervision.

\section{Behavioral Analysis of Reasoning Traces}
\label{sec:analysis}

We next  analyze the models' reasoning behaviors using executable counterfactuals.
Table~\ref{tab:reasoning-error-types} illustrates the three types of prototypical failure that we observe in the reasoning traces:

\begin{enumerate}[noitemsep,topsep=0pt,parsep=0pt,partopsep=0pt]
    \item Brute-force enumeration of all possible hidden-variable values.
    \item Assuming an arbitrary value for the hidden variable once the problem is considered too complex.
    \item Complicating the problem through unnecessary case splitting and circular analyses.
\end{enumerate}

Inspired by these observations, we evaluate each reasoning trace along two dimensions: \textit{planning} and \textit{execution}.
The planning score evaluates whether the three core cognitive skills of \cf reasoning---abduction, intervention, and prediction---are sequentially applied.
The execution score evaluates the correctness of mathematical computation and code simulation, a general skill that is not specific to \cf reasoning. 
Following prior work~\citep{sun2025omega}, we use o4-mini as the LLM judge to rate each dimension on a scale of 1 to 5, and defer other technical details, including the grading rubric in prompts, to Table~\ref{tab:evaluation_prompt_o4} in Appendix.
Figure~\ref{fig:code_cf-o4mini_analysis-7b} presents the results.

\begin{table}[t]
\setlength{\tabcolsep}{5pt}
\centering
    \begin{tabular}{C{0.45\textwidth} C{0.45\textwidth}}
    \toprule
    \textbf{Brute-Force Enumeration} & \textbf{Arbitrary Assumption} \\
    \midrule
    Since $r$ is not given, list $r$ and unroll loops:
    $r=0 \to \texttt{local\_sum}=1 \to \lfloor 1/3 \rfloor=0$;
    $r=1 \to \texttt{local\_sum}=3 \to \lfloor 3/3 \rfloor=1$;
    $r=2 \to \texttt{local\_sum}=5 \to \lfloor 5/3 \rfloor=1$;
    \ldots\ scan until $y=120$ fits at $x=15$.
    &
    Assume $r=3$, otherwise would be too complex to solve. One outer iter: $\texttt{local\_sum}=3+(3+1)=7 \to \lfloor 7/3 \rfloor=2$.
    For $x=12$: $y=12\cdot 2+12=36$. \\
    \midrule
    \multicolumn{2}{C{0.96\textwidth}}{\textbf{Unnecessary Case-Splitting}} \\
    \cmidrule(lr){1-2}
    \multicolumn{2}{C{0.96\textwidth}}{
    Recover $r$ from $x=15$, $y=120$ via inner-loop stops.
    Case 1: $r\le 0 \to$ inner never runs $\to y=x=15$.
    Case 2: $r>0 \to$ $\texttt{local\_sum}$ after step1 is $r$.
    Split 2A: stop after step1 if $r\ge 5r$.
    Split 2B: take step2 if $r<5r \to \texttt{local\_sum}=2r+1$.
    Also split by $2r+1$ vs. $5r$ ($<,=,>$) and by $(2r+1)\bmod 3\in\{0,1,2\}$;
    then branch on $q=\lfloor (2r+1)/3 \rfloor \in \{6,7,8\}$ 
    \ldots
    } \\
    \bottomrule
    \end{tabular}
\vspace{2mm}
\caption{Examples of three prototypical failure modes in model-generated reasoning traces.}
\label{tab:reasoning-error-types}
\end{table}

\para{Scaling model size improves computational accuracy, but not abduction skills.}
As shown in Figure~\ref{fig:code_cf-o4mini_analysis-7b}, across all four coding tasks, scaling up the size of Qwen2.5-Instruct models leads to consistent improvements in execution ratings, but \emph{not} in planning.
Instead, the 7B model consistently receives higher ratings for its abduction skills than 32B on 3/4 tasks, and scores even higher than the 72B variant on both \textit{if\_else-long} and \textit{while}.
This suggests that scaling up the size of LLMs
that are post-trained on general domains improves the final accuracy in a way that does not comply with the standard ``abduction-intervention-prediction'' strategy, thus resulting in poor \cf reasoning performance even with a large model size.

\begin{figure}[t]%
  \centering
  \includegraphics[width=\textwidth]{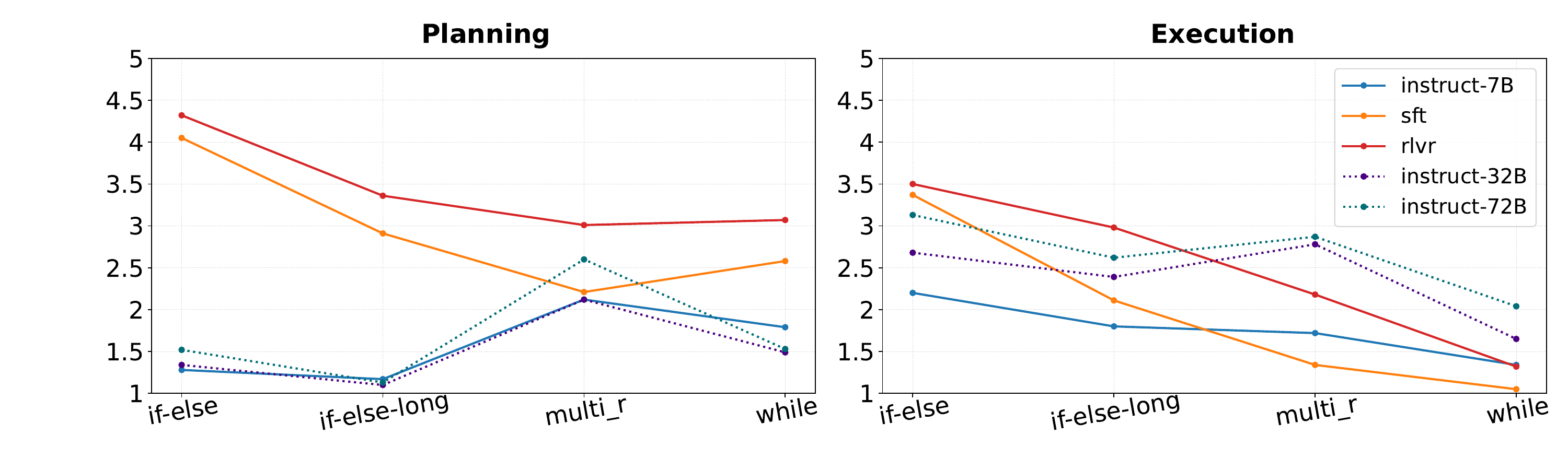}
  \vspace{-5mm}
  \caption{
  Evaluation results of the LLM-as-a-judge pipeline. For the responses generated by each model on each dataset, the evaluation objective is decoupled into ``planning'' (row 1; i.e., whether the ``abduction-intervention-prediction'' strategy is faithfully followed) and ``execution'' (row 2; i.e., whether the intermediate computations are correctly performed). 
  }
  \label{fig:code_cf-o4mini_analysis-7b}
  \vspace{-3mm}
\end{figure}

\para{SFT memorizes shallow abduction patterns that fail to generalize to complex problems.}
In Figure~\ref{fig:code_cf-o4mini_analysis-7b}, the planning scores of SFT models substantially drop in OOD tasks.
Our inspection of reasoning traces shows that 
when faced with OOD questions with increased complexity in completing the abduction step, 
SFT models tend to override the standard reasoning strategy, and instead revert to 
the prototypical failure modes discussed in Table~\ref{tab:reasoning-error-types}
in order to evade true \cf reasoning.

\para{RLVR generalizes \cf reasoning strategies, but is still bottlenecked by computational accuracy.}
As Figure~\ref{fig:code_cf-o4mini_analysis-7b} also reveals, RLVR models achieve the highest planning scores across all evaluation datasets,
demonstrating the generalizable \cf reasoning strategy that they learn to apply even in fundamentally OOD tasks.
On the other hand, 
the sharp decrease in execution scores on both \textit{multi\_r} and \textit{while}
also suggests that a major error type for RLVR is computational errors under the correct reasoning strategy.
Therefore, our framework identifies the asynchronism in learning \cf reasoning skills and general computational skills, and calls for future efforts into improving both skills simultaneously to build a strong \cf reasoning agent.

\section{Conclusion}
\noindent We address gaps in evaluating counterfactual reasoning in LLMs by decomposing the skill into core components by introducing an executable, code-based framework. Our setup builds dynamic testbeds that require the full abduction, action, prediction rollout and allows for precise control over logic and latent features. Using a template-based approach, we generate many structurally diverse functions to form counterfactual queries and to train smaller models that currently struggle on these tasks. We find that LLMs typically struggle at the abduction step, and this limitation is not resolved by increasing model's size as large scale models (up to 72B) also struggle with this. Our findings show that models trained with SFT transfer these skills in-domain code evaluations but significantly falter on OOD settings, whereas RL consistently induces them from code-only training and generalizes to novel control flows and natural-language counterfactual math. We corroborate this with qualitative case studies and \textit{LLM-as-a-Judge} evaluations. Beyond counterfactuals, the same framework enables flexible evaluation of other causal skills and can help pinpoint where current systems fall short.

\subsection*{Acknowledgements}
We are grateful to Abhinav Kumar, Daman Arora, Deema Alnuhait, Divyat Mahajan, Dylan Zhang, Junlin Yang, Kabir Ahuja, Lifan Yuan, Melanie Sclar, Nick Haber, Shivam Agarwal, Tobias Gerstenberg, Yanai Elazar, Yonatan Belinkov and members of Chicago Human+AI Lab for their valuable support and insightful discussions. This project is partly supported by NSF under award No. 2019897, No. 2126602, an award from the Sloan foundation, and an award from the Open Philanthropy foundation.

\bibliography{iclr2026_conference}
\bibliographystyle{iclr2026_conference}

\clearpage
\appendix

\onecolumn

\label{sec:ladder}
\section{Causal Ladder: Levels of Causal Reasoning}
\label{sec:causal_ladder}
The seminal work of \citet{pearl2009causality} breaks down causal reasoning in three progressively more advanced levels: \term{Associational} (level 1), \term{Interventional} (level 2), and \term{Counterfactual} (level 3):
\begin{itemize}
    \item \textbf{Associational level} concerns observational learning and forms causal hypothesis solely through observations, often interpreted as pattern matching. 
This mirrors how most machine learning models learn from input features and corresponding labels. This form of learning suffers from potential confounding and selection bias as one cannot perform interventions to identify the underlying causal structure.%
\item \textbf{Interventional learning (level 2)} requires learning through interventions, mirroring how humans typically learn by taking actions and observing the outcomes. 
While this type of learning might appear to be causal, due to the overall noise in the system which might be changing, identifying whether the observed outcome was solely due to the action performed becomes challenging.%
\item \textbf{Counterfactual reasoning (level 3)} is the highest form of causal reasoning on the causal ladder. 
It helps in disentangling the effect of other factors in the system, to identify the outcome had the original action been  different. However this requires stronger, unit-level structural assumptions, many counterfactuals are not identifiable from data without modeling and this form of reasoning is typically sensitive to model misspecification and ``cross-world'' assumptions.%
\end{itemize}

\begin{table*}[ht]
  \setlength{\tabcolsep}{5pt} %
  \renewcommand{\arraystretch}{1.2} %
  \centering

  \newlength{\tblavail}
  \setlength{\tblavail}{\dimexpr\linewidth - 10\tabcolsep\relax} %

  {\fontsize{9}{11}\selectfont
      \begin{tabular}{@{}%
          C{0.06\tblavail} %
          C{0.17\tblavail} %
          C{0.16\tblavail} %
          C{0.11\tblavail} %
          C{0.25\tblavail} %
          C{0.25\tblavail} %
      @{}}
        \toprule
        \textbf{Level} & \textbf{Concept} & \textbf{Expression} & \textbf{Activity} & \textbf{Question} & \textbf{Example} \\
        \midrule
        
        I &
        Association / Correlation &
        $P(y \mid x)$ &
        Seeing / Observing &
        How does seeing x change my belief in y? &
        Would the grass be dry if we found the sprinkler off? \\
        \midrule
        
        II &
        Intervention / Hypotheticals &
        $P(y \mid \mathrm{do}(x))$ &
        Doing &
        Would y happen if I did x? &
        Would the grass be dry if we made sure that the sprinkler was off? \\
        \midrule
        
        III &
        Counterfactuals &
        $P(y_x \mid x', y')$ &
        Imagining &
        Would y have happened instead of $y'$, if I had done $x$ instead of $x'$? / What would have happened if I had done $x$, given that doing $x'$ led to $y'$? &
        Would the grass have been dry if the sprinkler had been off, given that the grass is wet and the sprinkler on? \\
        \bottomrule
      \end{tabular}
  }
  \caption{
    Definition of the causal ladder proposed by Pearl~\citep{pearl2009causality}, where $\{x, x'\}$ denote candidate causes, $\{y, y'\}$ denote candidate effects, and $P$ denotes the probability of an event. Notably, Interventions and Hypotheticals are different names of the same reasoning paradigm, which only thinks about changes that lie in the \textbf{future}~\citep{gerstenberg2022would}. Counterfactuals differs from them by thinking about changes that lie in the observed outcome or \textbf{past}.
  }
  \label{tab:ladder_definitions}
\end{table*}

\section{\cf VS \interv: Examples of GSM-based Tasks}
\begin{table}[H]
\centering
\small
\begin{tabularx}{\linewidth}{@{}l >{\raggedright\arraybackslash}X r@{}}
\toprule
\textbf{Setting} & \textbf{GSM Problem} & \textbf{Answer} \\
\midrule
\textbf{Counterfactual} &
Ravi is organizing an office lunch. Every catering tray is priced at \$68.
There is also a \textcolor{red}{per-catering tray service fee}. A discount of 14\% is applied
to the items subtotal (before any fees). For 6 catering trays, the total shown
is \$353.88. If instead 11 catering trays were ordered, with all else unchanged,
what amount would be shown? & \$648.78 \\
\addlinespace[0.25em]
\textbf{Interventional} &
Ravi is organizing an office lunch. Every catering tray is priced at \$68.
There is also a \textcolor{blue}{per-catering tray service fee of \$0.50}. A discount of 14\% is applied
to the items subtotal (before any fees). For 6 catering trays, the total shown
is \$353.88. If instead 11 catering trays were ordered, with all else unchanged,
what amount would be shown? & \$648.78 \\
\bottomrule
\end{tabularx}
\caption{Two GSM-style instances derived via the dependency-graph approach inspired by \citet{ye2024physicslanguagemodels21}. The first row is a \emph{counterfactual} with a hidden latent variable (highlighted) that must be inferred; the second row is the corresponding \emph{interventional} instance with the fee (hidden latent variable) revealed. }
\label{tab:cf-example}
\end{table}

\section{\cf VS \interv: Performance on GSM-based Tasks}

\begin{table}[H]
\centering
\small
\begin{tabular}{lcc}
\toprule
\textbf{Model} & \textbf{GSM-Interventional} & \textbf{GSM-Counterfactual} \\
\midrule
\multicolumn{3}{l}{\emph{}} \\
Qwen2.5-1.5B-Instruct            & 
    \textbf{18.4} & 9.0 \\
Qwen2.5-3B-Instruct              
    & \textbf{40.3} & 22.0 \\
Qwen2.5-7B-Instruct              
    & \textbf{60.4} & 39.0 \\
Qwen2.5-32B-Instruct             
    & \textbf{88.5} & 73.1 \\
Qwen2.5-72B-Instruct             
    & \textbf{79.7} & 73.1 \\
Llama-3.3-70B-Instruct           
    & \textbf{95.1} & 82.2 \\
GPT-4o                           
    & \textbf{93.1} & 68.6 \\
DeepSeek-R1-Distill-Qwen-32B     
    & \textbf{75.3} & 60.7 \\
\bottomrule
\end{tabular}
\caption{
Performance comparison on GSM-based \cf and \interv tasks for various models, where the latter is still consistently and significantly higher than the former, echoing prior observation in Figure~\ref{fig:code_cf_vs_interv} for code-based tasks.
}
\label{tab:gsm_cf_vs_interv}
\end{table}

\section{When Counterfactuals and Interventionals Conflate}
\label{sec:interv_same_cf}

\citet{wu-etal-2024-reasoning} analyze GPT-4 under altered premises; 
since their tasks contain no latent variables, intervention and counterfactual queries coincide, so the reported failures does not probe abductive backtracking effectively. For example 
one of their evaluated tasks in arithmetic, switching the base from 10 to 9 is simply \(\operatorname{do}(\text{base}=9)\): \(27_{10}+62_{10}=89_{10}\) but \(27_{9}+62_{9}=(100)_9\); no latent state needs to be inferred. Consequently, these setups don't diagnose whether a model can perform abduction. 
Using Fig.~\ref{fig:cf_vs_interv_pipeline} as an comparison, the base is $x$ and there is no $r$ in this example. 
Similar approaches are also adopted in previous works \citep{li2024promptinglargelanguagemodels, nguyen2024llmsgeneratingevaluatingcounterfactuals, paranjape-etal-2022-retrieval, wu-etal-2021-polyjuice, madaan2021generatecounterfactualscontrolledcounterfactual, ye-etal-2021-connecting, joshi2022investigationineffectivenesscounterfactuallyaugmented, vashishtha2023evaluatingmitigatinggenderbiases}.
Most of these works use counterfactuals for robustness, debiasing and other purposes. They operate in fully observed settings without latent variability, where the query effectively reduces to an intervention. In contrast, our evaluation targets cases that require abduction, testing whether LLMs can execute the full Abduction-Action-Prediction rollout.

\section{Additional Related Work}
\paragraph{Causality and LLMs.} Recently a lot of work has focused on how effectively LLMs can be used as domain priors for discovering causal relationship between different real world entities ~\citep{kiciman2023causal, ban2023query, long2023causal, willig2023probing, vashishtha2025causalorderkeyleveraging}.
Furthermore, some efforts have also focused on improving LLM's causal reasoning via training on synthetic data \citep{vashishtha2025teachingtransformerscausalreasoning}, or by testing different Chain-of-Thought (CoT) based methods \citep{jin2024cladderassessingcausalreasoning}. Works like \cite{jin2023can, jin2024cladderassessingcausalreasoning} underline the current limitations of language models' causal reasoning in synthetic and formal settings across different types of reasoning including counterfactual reasoning.

\paragraph{Using Counterfactuals for NLP tasks:} Past work has also been focusing on improving robustness in NLP tasks such as debiasing for gender based associations~\citep{wu-etal-2024-reasoning,paranjape-etal-2022-retrieval, wu-etal-2021-polyjuice, madaan2021generatecounterfactualscontrolledcounterfactual, ye-etal-2021-connecting, joshi2022investigationineffectivenesscounterfactuallyaugmented, vashishtha2023evaluatingmitigatinggenderbiases}, story generation~\citep{qin-etal-2019-counterfactual}, fictional complex reasoning ~\citep{ahuja2025findingflawedfictionsevaluating}, and improving efficiency of  reasoning trace ~\citep{lu2025retrosearchexploringuntakenpaths}. Recent work uses \textit{abduction-action-prediction} based promnpting strategy for accurate failure attribution in multi-agent systems for tasks like debugging, showing promising improvement \citep{west2025abductactpredictscaffolding}.
These works have used counterfactuals as a way to improve robustness in language models, and test reasoning abilities, while following a simplified interpretation of counterfactual reasoning \citep{pearl2002reasoning}.

\paragraph{RL vs SFT:} Past studies have explored how the training paradigms of SFT and RL based training differ, which guide our training design setup. \citet{kirk2024understandingeffectsrlhfllm} shows how Reinforcement Learning from Human Feedback (RLHF), generalizes better then SFT under distribution shift from train set, however results in lack of diversity. \citet{chu2025sftmemorizesrlgeneralizes} showed how RL trained on outcome based reward generalizes better across both text and visual tasks, while SFT memorizes the task leading to lack of generalization. However the work emphasises the importance of SFT before RL for effective training. \citet{wu2025generalizationsftreinforcementlearning} shows how standard SFT's lack of generalization is due to gradients encoding problematic reward leading to lack of generalization.  

We take inspiration from cognitive science literature \citep{gerstenberg2022would} to design our program-based analysis in order to evaluate the core cognitive skills required for counterfactual inference. To build math-based generalization tests we build upon the framework of \citet{ye2024physicslanguagemodels21} to generate counterfactual variant of grade school level math problems following a dependency graph based approach. Past work uses programs as world models and simulations, including concept learning from programs \citep{lake2015humanlevel} and code driven or physics based simulators \citep{cobbe2020leveragingproceduralgenerationbenchmark, freeman2021braxdifferentiablephysics} showing the potential of graph for this, which we leverage for our work.

\section{Meta Templates: Structural Placeholder Description}

\begin{table}[ht]
\centering
\small
\begin{tabularx}{\linewidth}{@{}lXX@{}}
\toprule
\textbf{Placeholder} & \textbf{What it controls} & \textbf{Code/line type inserted} \\
\midrule
\texttt{\{function\_name\}} & Name of the generated function. & Identifier used in \texttt{def} header (snake\_case). \\
\texttt{\{min\_r\}}, \texttt{\{max\_r\}} & Bounds for the random draw \texttt{r}. & Integer literals or simple expressions inside \texttt{random.randint(a,b)}. \\
\texttt{\{preprocessing\_block\}} & Optional setup before branching. & One or more Python statements (e.g., assignments, helper calls). \\
\texttt{\{condition\_type\}} & The top-level \texttt{if} condition. & Boolean expression (comparisons, logical ops). \\
\texttt{\{if\_branch\_content\}} & Body when \texttt{if} is true. & Indented suite: one or more Python statements. \\
\texttt{\{elif\_block\}} & Optional middle branch. & Either empty, or \texttt{elif <boolean expr>:} + indented suite. \\
\texttt{\{else\_branch\_content\}} & Body when previous conditions are false. & Indented suite: one or more Python statements. \\
\texttt{\{return\_expression\}} & Value the function returns. & Expression used in \texttt{return} (identifier, arithmetic expr, tuple, etc.). \\
\bottomrule
\end{tabularx}
\caption{Placeholders, roles, and expected code/line types for the if--else meta-template.}
\label{tab:cf_prompt_meta_template}
\end{table}

\section{Deduplication and Verification of Code Functions}
\label{sec:deduplication}
We validate each function by executing it on a small, randomly generated verification set to ensure it runs without errors.. We also parse the code into Python's Abstract Syntax Tree (AST) to confirm that it compiles without syntax errors. For computing the similarity we convert each generated function into a \textit{structural fingerprint} by counting key elements (if-statements, assignments, operators) and analyzing the overall code pattern. To analyze patterns, it walks through the code structure and identifies sequences like ``preprocessing $\rightarrow$ condition check $\rightarrow$ branch calculations $\rightarrow$ return result''. It then compares these fingerprints numerically: if two functions have similar counts of each element type and follow the same logical flow pattern, they get a high similarity score $s \in [0,1]$. Functions with identical structure and execution sequence get a score of $s = 1.0$, while completely different functions score near $s = 0$. Based on manual analysis, we set the threshold at 0.8. This helps identify when the generation process is creating duplicate or overly similar functions that should be filtered out to maintain training data diversity.

\section{Counterfactual Reasoning Prompts for Code Functions}\label{sec:appendix_cf_prompts}

\subsection{While}

\begin{table}[H]
    \centering
    \footnotesize
    \begin{tabular}{m{12cm}}
        \toprule
        \textit{You are a language model that reasons about code without using any external execution environment. Do not simply repeat the prompt. Instead, analyze the Python function below, provide step-by-step reasoning, and answer the counterfactual question.} \\
        \\
        \textit{\textbf{Python function:}} \\
        \begin{minipage}{\linewidth}
        \vspace{2pt}
\begin{lstlisting}[language=Python]
def generated_func_997660_100(x, r):
    
    primary_sum = 0
    secondary_sum = 0
    counter = 0
    
    while counter < x:
        primary_sum += r + counter
        secondary_sum += counter * 2
        
        if primary_sum > secondary_sum:
            primary_sum -= 5
        
        counter += 1
    
    return (primary_sum + secondary_sum) // 5
\end{lstlisting}
        \vspace{-6pt}
        \end{minipage} \\
        \\
        \textit{\textbf{Observed call:}} \\
        \textit{When this function was called with input \(x = 10\), it produced the output \(y = 36\).} \\
        \\
        \textit{\textbf{Counterfactual query:}} \\
        \textit{If instead of \(x = 10\), we had called this function with a different input value of \(x = 8\) while keeping everything else unchanged, what could the output \(y\) have been? Let's think step by step to get the answer. 
        } \\
        \\
        \textit{\textbf{Required answer format:}} \\
        \verb|\boxed{ans1, ans2, ans3}| \\
        \bottomrule
    \end{tabular}
    \vspace{2mm}
    \caption{
    Counterfactual prompt example for \textit{While} dataset.
    }
    \label{tab:counterfactual_prompt-diverse_while}
\end{table}

\subsection{Multi\_r}

\begin{table}[H]
    \centering
    \footnotesize
    \begin{tabular}{m{12cm}}
        \toprule
        \textit{You are a language model that reasons about code without using any external execution environment.
        Do not simply repeat the prompt. Instead, analyze the Python function below, provide step-by-step reasoning, and answer the counterfactual question.} \\
        \\
        \textit{\textbf{Python function:}} \\
        \begin{minipage}{\linewidth}
        \vspace{2pt}
\begin{lstlisting}[language=Python]
import random

def generated_func_1136(x, r1, r2, r3):
    
    prep = x * (r2 + r3)
    
    if x == r1:
        result = x * r3
        for i in range(2):
            pass
        result = result = x + r2
    else:
        result = x - r2
        for j in range(6):
            pass
        result = result = x + r1
    
    result = result * (r1 + r2 * r3)
    return result
\end{lstlisting}
        \vspace{-6pt}
        \end{minipage} \\
        \\
        \textit{\textbf{Observed call:}} \\
        \textit{When this function was called with input \(x = 16\), it produced the output \(y = 3640\).} \\
        \\
        \textit{\textbf{Counterfactual query:}} \\
        \textit{
        If instead of \(x = 16\), we had called this function with a different input value of \(x = 18\) while keeping everything else unchanged, what could the output \(y\) have been? 
        Let's think step by step to get the answer. 
        } \\
        \\
        \textit{\textbf{Required answer format:}} \\
        \verb|\boxed{ans1, ans2, ans3}| \\
        \bottomrule
    \end{tabular}
    \vspace{2mm}
    \caption{
    Counterfactual prompt example for \textit{Multi\_r} dataset
    }
    \label{tab:counterfactual_prompt-multi_r}
\end{table}

\subsection{If\_else-long}

\begin{table}[H]
    \centering
    \footnotesize
    \begin{tabular}{m{12cm}}
        \toprule
        \textit{You are a language model that reasons about code without using any external execution environment. Do not simply repeat the prompt. Instead, analyze the Python function below, provide step-by-step reasoning, and answer the counterfactual question.} \\
        \\
        \textit{\textbf{Python function:}} \\
        \begin{minipage}{\linewidth}
        \vspace{2pt}
\begin{lstlisting}[language=Python]

def generated_func_1194(x, r):
    alt4 = 10
    final2 = 1
    final3 = 0
    final4 = 4
    temp1 = 3
    temp2 = 3
    temp3 = 2
    r = abs(r)
    
    if r > 9:
        temp1 = (x %
        if (r %
            
            if temp1 < 5:
                
                if (temp3 * x) < r:
                    final4 = (temp3 * x) + 2
                    result = final4 + x
                else:
                    alt4 = x - temp3
                    result = alt4 + r
            else:
                final3 = temp2 + r
                result = final3 - x
        else:
            final2 = (temp1 ** 5) * r
            result = final2 * r
    else:
        else_val = (r ** 4) * x
        result = else_val + r
    
    return result %
\end{lstlisting}
        \vspace{-6pt}
        \end{minipage} \\
        \\
        \textit{\textbf{Observed call:}} \\
        \textit{When this function was called with input \(x = 18\), it produced the output \(y = 4\).} \\
        \\
        \textit{\textbf{Counterfactual query:}} \\
        \textit{If instead of \(x = 18\), we had called this function with a different input value of \(x = 20\) while keeping everything else unchanged, what could the output \(y\) have been? Let's think step by step to get the answer.
        } \\
        \\
        \textit{\textbf{Required answer format:}} \\
        \verb|\boxed{ans1, ans2, ans3}| \\
        \bottomrule
    \end{tabular}
    \vspace{2mm}
    \caption{
    Counterfactual prompt example for \textit{If\_else-long} dataset.
    }
    \label{tab:counterfactual_prompt-diverse_ifelselong}
\end{table}

\section{Interventional Reasoning Prompts for Code Functions}
\begin{table}[H]
    \centering
    \footnotesize
    \begin{tabular}{m{12cm}}
        \toprule
        \textit{You are a language model that reasons about code without using any external execution environment.
        Do not simply repeat the prompt. Instead, analyze the Python function below, provide step-by-step reasoning, and answer the \textbf{interventional} question.} \\
        \\
        \textit{\textbf{Python function:}} \\
        \begin{minipage}{\linewidth}
        \vspace{2pt}
\begin{lstlisting}[language=Python]
def generated_func_1273(x, r1, r2, r3):
    
    prep = x + (r2 - r3)
    
    if x != r1:
        result = x + r2
        for i in range(6):
            pass
        result = result = x * r1
    else:
        result = x + r3
        for j in range(2):
            pass
        result = result = x + r3
    
    result = result + (r1 - r2 - r3)
    return result
\end{lstlisting}
        \vspace{-6pt}
        \end{minipage} \\
        \\
        \textit{\textbf{Observed call:}} \\
        \textit{When this function was called with inputs \(x = 18\), \(r_1 = 20\), \(r_2 = 5\), and \(r_3 = 17\), it produced the output \(y = 358\).} \\
        \\
        \textit{\textbf{Interventional query:}} \\
        \textit{
        If instead of \(x = 18\), we had called this function with \(x = 20\) while keeping \(r_1 = 20\), \(r_2 = 5\), and \(r_3 = 17\) unchanged, what \emph{could} the output \(y\) have been? Let's think step by step to get the answer.
        } \\
        \\
        \textit{\textbf{Required answer format:}} \\
        \verb|\boxed{ans1, ans2, ans3}| \\
        \bottomrule
    \end{tabular}
    \vspace{2mm}
    \caption{
    Interventional prompt example for \textit{Multi\_r} dataset.
    We omit the \interv examples of the remaining three code-based datasets, as their prompt templates are mostly similar to this one, and the examples of python functions are already displayed in Appendix~\ref{sec:appendix_cf_prompts}
    }
    \label{tab:interventional_prompt_multir}
\end{table}

\section{Prompt Template for LLM-as-a-judge analysis}
\begin{table}[H]
    \centering
    \footnotesize
    \begin{tabular}{m{12cm}}
        \toprule
        \textit{You are presented with a counterfactual reasoning question about a code function, along with a sample solution. Your task is to carefully analyze this solution and rate how it performs in terms of planning and execution, on a scale from 1 to 5.} \\
        \\
        \textit{\textbf{Criteria for rating planning:}} \\
        \textit{\textbf{5} -- The solution adopts a perfect plan for all such counterfactual questions with two stages: (1) \emph{Backward Reasoning with Original Data}: determine the value(s) of the unknown variable \texttt{r} by setting up a mathematical equation based on the arithmetic operations performed on \texttt{r} and the original input \texttt{x} within the code path that produced the original output \texttt{y}. (2) \emph{Forward Reasoning with Counterfactual Data}: use the value(s) of \texttt{r} found in the previous step to determine the new output(s) based on the counterfactual input.} \\
        \textit{\textbf{3} -- The solution shows awareness of first finding values of \texttt{r} from the original \texttt{x} and \texttt{y}, and then computing the new outputs using the same \texttt{r}, \emph{but it does not follow this Backward-then-Forward plan faithfully or decisively}. For instance, it hesitates about solvability without an explicit \texttt{r}, or resorts to brute-force enumeration without persisting in the desired plan.} \\
        \textit{\textbf{1} -- The solution does not align with the Backward-then-Forward plan at all (e.g., starts with brute-force enumeration of \texttt{r} without using the given \texttt{x}, \texttt{y} to determine \texttt{r} smartly).} \\
        \\
        \textit{\textbf{Criteria for rating execution:}} \\
        \textit{Score execution based on whether the solution follows the code-simulation paths and performs step-by-step numerical computations faithfully and correctly. More simulation/computation mistakes $\rightarrow$ lower execution score.} \\
        \\
        \textit{\textbf{Question:}} \\
        \texttt{\{prompt\}} \\
        \\
        \textit{\textbf{Ground Truth Answers:}} \\
        \texttt{\{ground\_truth\}} \\
        \\
        \textit{\textbf{Solution:}} \\
        \texttt{\{response\}} \\
        \\
        \textit{\textbf{Required response format (JSON):}} \\
        \begin{minipage}{\linewidth}
        \vspace{2pt}
\begin{lstlisting}[language=json]
{
    "planning": [1|2|3|4|5],
    "planning_explanation": "first briefly describe the planning or strategy this solution adopts, and then explain why you gave this planning score",
    "execution": [1|2|3|4|5],
    "execution_explanation": "brief explanation of why you gave this execution score"
}
\end{lstlisting}
        \vspace{-6pt}
        \end{minipage} \\
        \bottomrule
    \end{tabular}
    \vspace{2mm}
    \caption{
    The prompt template for LLM-as-a-judge analyses in \S\ref{sec:analysis}, with detailed evaluation rubrics for both \textbf{planning} and \textbf{execution} scores.
    }
    \label{tab:evaluation_prompt_o4}
\end{table}

\section{Training and Evaluation Details}

\subsection{SFT Training Setups}\label{sec:sft_config}

\paragraph{Reasoning Trace Generation.} 
Our training dataset is built upon 5500 code-based \cf reasoning prompts that only involve \textit{if\_else} logic.
We leverage \textit{DeepSeek-Distilled-Qwen-32B-Instruct}~\cite{guo2025deepseek} to annotate the reasoning traces for these prompts through rejection sampling~\cite{li2025naturalthoughts}.
Specifically, for each prompt, we sample multiple responses until the model reaches the correct final answer or a budget of $N = 8$ is reached.
The resulting training set achieves a final F1 score of 95.9 and exact match score of 88.9, ensuring the correctness of SFT training signals.

\paragraph{Training Configurations.}
Throughout this work, we follow prior practice~\citep{ye2025limo} by performing full-parameter SFT training using the LlamaFactory framework~\citep{zheng-etal-2024-llamafactory}. 
All the SFT experiments are carried out using four NVIDIA H100 GPUs, with DeepSpeed Zero-3~\citep{10.1145/3394486.3406703}, FlashAttention-V2~\citep{dao2022flashattention}, and Liger Kernel~\citep{hsu2025ligerkernel} enabled to improve time and memory efficiency. 
The key training hyperparameters are shown below:

\begin{verbatim}
--cutoff_len 16384
--num_train_epochs 2
--bf16 True
--optim adamw_torch
--lr_scheduler_type cosine
--learning_rate 5e-05
--warmup_ratio 0.05
--weight_decay 0.0
--per_device_train_batch_size 1
--gradient_accumulation_steps 4
--seed 42
\end{verbatim}

\subsection{RL Training Setups}\label{sec:rl_config}

Throughout this work, we follow prior practice~\citep{luo2025through, sun2025omega} by performing full-parameter RLVR training with the verl~\citep{sheng2024hybridflow} framework.
We use four NVIDIA H100 GPUs for training 1.5B and 3B models, and eight H100 GPUs for 7B models.
We adopt an effective prompt batch size of 16, a rollout batch size of 24, a prompt length limit of 512, a response length limit of 2000, a sampling temperature of 1.0, a coefficient of $10^{-3}$ for low-variance KL auxiliary loss, and a total of 1500 training steps. The reward is the simple exact match score.

\subsection{Evaluation Setups}\label{sec:eval_config}

\paragraph{Dataset Statistics.}
The in-distribution (ID) \textit{If\_else} evaluation dataset contains 500 examples, while the three out-of-distribution (OOD) evaluation datasets, \textit{If\_else-long}, \textit{Multi\_r} and \textit{While}, contain 480, 575 and 480 examples respectively.

\paragraph{Evaluation Protocol.}

Throughout this work, we use the vLLM framework~\citep{kwon2023efficient} for efficient evaluation. 
Specifically, we follow prior practice~\citep{luo2025through, guo2025deepseek} by using sampling with a temperature of 0.6, a top-p of 0.95, and a maximum of 16000 generated tokens to generate $k = 3$ responses per question. For each evaluation dataset, we report the average accuracy over $k$ responses (i.e., avg@k) to reduce the variance in performance statistics.

\end{document}